# Stepwise Acquisition of Dialogue Act Through Human-Robot Interaction

Akane Matsushima
*School of Science and Technology*
*Kyoto Institute of Technology*
Kyoto, Japan
matsu@ii.is.kit.ac.jp

Ryosuke Kanajiri
*School of Science and Technology*
*Kyoto Institute of Technology*
Kyoto, Japan
ryosuke0947@gmail.com

Yusuke Hattori
*Graduate School of Science and Technology*
*Kyoto Institute of Technology*
Kyoto, Japan
yusu11x1225@gmail.com

Chie Fukada
*Faculty of Arts and Sciences*
*Kyoto Institute of Technology*
Kyoto, Japan
chieft@kit.ac.jp

Natsuki Oka
*Faculty of Information and Human Sciences*
*Kyoto Institute of Technology*
Kyoto, Japan
nat@kit.ac.jp

*Abstract*—A dialogue act (DA) represents the meaning of an utterance at the illocutionary force level (Austin 1962) such as a question, a request, and a greeting. Since DAs take charge of the most fundamental part of communication, we believe that the elucidation of DA learning mechanism is important for cognitive science and artificial intelligence, especially for cognitive robotics. The purpose of this study is to verify that scaffolding takes place when a human teaches a robot, and to let a robot learn to estimate DAs and to make a response based on them step by step utilizing scaffolding provided by a human. To realize that, it is necessary for the robot to detect changes in utterance and rewards given by the partner and continue learning accordingly. Experimental results demonstrated that participants who continued interaction for a sufficiently long time often gave scaffolding for the robot. Although the number of experiments is still insufficient to obtain a definite conclusion, we observed that 1) the robot quickly learned to respond to DAs in most cases if the participants only spoke utterances that match the situation, 2) in the case of participants who built scaffolding differently from what we assumed, learning did not proceed quickly, and 3) the robot could learn to estimate DAs almost exactly if the participants kept interaction for a sufficiently long time even if the scaffolding was unexpected.

## I. Introduction

A dialogue act (DA) represents the meaning of an utterance at the illocutionary force level [1], for example, questions, requests, and greetings. Understanding DAs is the first step toward intention comprehension and is essential for the agent to properly respond to people, and hence many studies on DA classification have been done. Although many classification methods have been tried [2, 3], recently, the technique using deep learning shows good performance [4–6] due to the rapid progress of the technology.

On the other hand, from the viewpoint of requiring less computation, DA classification based solely on function words was reported [7]. Though function words such as articles, prepositions, determiners etc. are not important for information retrieval purposes, they contain sufficient information for DA classification. For example, questions often have distinguishing features such as an interrogative (what, who) or an auxiliary verb (can, is) at the beginning.

Compared to the above-mentioned conventional research, the problem setting of this research differs in the following four points [8]:

1. While previous research dealt with supervised learning, this research deals with learning not from labeled examples but from rewards[1].

2. In conventional research, a lot of examples were batch processed, but in this research, a robot processes relatively few examples incrementally.

3. Conventional research aimed at the classification of DAs, but the purpose of this research is to enable a robot to respond appropriately according to DAs.

4. The way words and rewards given to learners varies by scaffolding.

Because the problem to solve is more difficult due to the difference in the settings in 1 to 4 above, we adopt a simple approach to estimate DAs only from function words. Furthermore, as we deal with interactions in Japanese in this research, we can estimate DAs almost exactly by focusing only on sentence-final particles which are one type of function words. This is because, in Japanese, the mental attitude of a speaker is usually expressed by sentence-final particles. For example, the question which is one of DAs is represented by a sentence-final particle *ka*.

---

[1] Although this paper focuses on reinforcement learning (RL), there is no intention to compare the learning efficiency between RL and supervised learning. In the future, we plan to study a model that combines both.

Scaffolding is a process that enables a child or novice to solve a problem, carry out a task or achieve a goal which would be beyond his unassisted efforts [9]. In this paper, we consider changes provided by a teacher to help a learner as scaffolding. As an example of scaffolding, Roy, Frank, and Roy showed that caregivers gradually decreased the length of their utterances containing a particular word type up to the moment of birth of that word, and then gradually increased complexity [10]. We assume that scaffolding will take place even when a person teaches a robot, not a child.

Recently, Qureshi, Nakamura, Yoshikawa, and Ishiguro applied deep Q-learning to the problem of a robot learning social actions (wait, look towards human, wave hand and handshake) through human-robot interaction [11]. They reported that the robot interpreted the human behavior by intention depicting factors (e.g., human body language, walking trajectory or any ongoing activity, etc.). Hermann et al. presented an agent that learns to interpret language in a simulated 3D environment where it is rewarded for the successful execution of written instructions [12]. They trained a single agent to execute phrasal instructions pertaining to multiple tasks by employing a curriculum training regime. Both [11] and [12] proposed a learning system based on reinforcement learning as in this study. [12] used curriculum learning, which can be regarded as a kind of scaffolding. The uniqueness of this study is that scaffolds were not prepared carefully by researchers, but were provided naturally by participants.

The purpose of this study is to let a robot learn to estimate DAs and to make a response based on them utilizing scaffolds. In the following sections, we will report on the setting of the first experiment, the first learning model of the robot, and the result of the experiment. Then we will discuss the limitation of the learning model and propose a revised model, which lets the robot learn step by step utilizing scaffolds provided by a human. Finally, the result of the second experiment will be described. Although the summary of the first experiment, the first learning model, and the result of the first experiment has been reported in a short paper [8], these are described fully in Section II because we could not explain the detail of the learning model, and only the small part of the result was shown in [8].

## II. DA Acquisition Through Human-Robot Interaction

### A. Setting in Experiment 1

In this section, we will describe the experimental setting of the first experiment. The human-robot interaction is performed in the following procedure [8]:

1. A participant puts one of the fruits (an apple or a banana) in front of a robot. The robot recognizes the fruit by its camera[2].

2. The participant speaks to the robot. Participant's utterances are limited to a combination of three function words ("*yo/ne/ka*") and three content words ("apple/banana/looks tasty"), for example, *ringo da yo*. 'This is an apple.'; *oishi sou da ne*. 'It looks tasty, isn't it?'; *banana desu ka?* 'Is this a banana?'

3. The robot recognizes the speech[3] and makes a response. Expression from the robot is basically a movement of the neck (a nod or a head shake), and in addition to this, the robot sometimes speaks *nee*[4]. 'That's right.'

4. The participant evaluates the response by pushing one of the + or − buttons. Participants were not instructed on what factors of the response should be evaluated.

See Fig. 1 for the experimental setting.

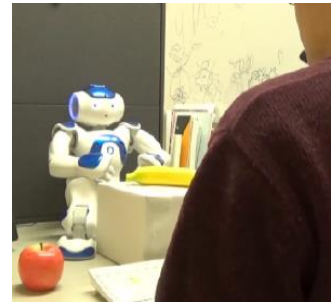

Fig. 1. Interaction between a human and a robot in Experiment 1.

Instructions for the participants before the experiment are as follows [8]:

- The robot does not know anything about the names of the fruits and the meaning of the sentence-final particles.

- Continue the interaction with the robot until you judge that the robot has completed the learning of all the following: the names of the fruits, the meaning of "looks tasty", the meaning of the sentence-final particles *yo*, *ne* and *ka*, and the appropriate usage of *nee*.

Under the above setting, the DA represented by each sentence-final particle and the appropriate response thereto are as follows:

*yo*: The speaker intends to teach the robot the name of the fruit or that it looks delicious. Hence the robot should memorize it and express the execution of internal processing by a nod.

*ne*: The speaker believes that the robot is also thinking that the fruit is there or it looks delicious and asks for the robot to agree. The speaker also wants to sympathize with the robot each other. The robot should confirm the presence of the fruit or its deliciousness based on its knowledge and should express it with a nod. The Robot should also enhance

---

[2] The accuracy of image recognition was 89%, but since the experimenter at the invisible position corrected the errors on the spot, the recognition result received by the learning system was 100% correct.

[3] In the case where the participant faltered or rephrased, erroneous recognition occurred, and the accuracy was 86%. Also, in this case, the experimenter at the invisible position corrected the errors on the spot.

[4] *Nee* is a prolonged version of *ne*, which is often used as a response to utterances ending with *ne*.

sympathy with the speaker and express it with the utterance *nee*.

*ka*: The speaker intends to ask a question in order to check the learning state of the robot. The robot thus should respond to the question by nodding or shaking its head based on the learned knowledge.

### B. Computational Model of DA Learning

We designed an agent which learns to recognize DAs through interaction described in the previous section. As shown in Fig. 2, it is designed to switch internal information processing blocks (light yellow squares in the drawing) by reinforcement learning (RL) blocks (open squares) according to the function words in human speech.

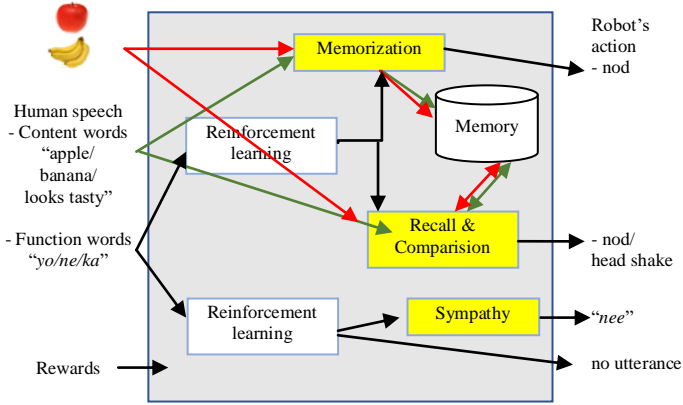

Fig. 2. Structure of the DA learning system

State $s$ in RL blocks consisted of the function words in human speech, and action $a$ in the upper RL block is the selection of the internal processing, i.e., 'memorization' or 'recall & comparison', and action $a$ in the lower RL block is either sympathetic or no utterance. For simplicity, we assumed that the rewards are given every time without fail, and excluded delayed rewards, which simplified the action value update as follows:

$$Q(s,a) \leftarrow Q(s,a) + \alpha(r - Q(s,a)) \quad (1)$$

where $Q(s,a)$ is an action-value function, that is, the value of taking action $a$ in state $s$, $\alpha$ is the learning rate, and $r$ is the reward. $\alpha$ was set to 0.1 and $r$ was either $+1$ or $-1$. The action $a$ was selected with probability

$$\frac{e^{Q(a)/\tau}}{\sum_{b=1}^{n} e^{Q(b)/\tau}} \quad (2)$$

where $\tau$ is a positive parameter called the temperature, $n$ is the total number of actions, and $b$ is one of the actions. $\tau$ was set to 0.16.

We also implemented the memorization and recalling process of content words as reinforcement learning (RL). In this case, state $s$ in RL was the recognition result of fruits, and action $a$ was one of the content words, "apple", "banana", or "looks tasty". If memorization was selected, action-value function $Q(s,a)$ was updated according to equation (1). For example, when a participant put an apple and said *ringo da yo*.

'This is an apple.', and if memorization was selected and a positive reward was given, $Q(apple, "apple")$ increased. However, when the robot shook its head, the reward for that action was interpreted with reversed polarity. In order to answer a question such as "Is this a banana?", it is necessary to compare the contents asked and the contents recalled from memory. If the recall probability, equation (2), exceeds a threshold, it was considered to be verified. Note that what the content words refer to is expressed as action values, which makes the learning robust, that is, it is safe to have noisy input.

Learning in this model involves the following two difficulties:

1. The robot learns not only its actions but also its internal processing, but it cannot be seen from the outside. Participants give rewards based solely on observable responses of the robot. Consequently, an appropriate reward is not always given.

2. In the robot, multiple learning proceeds in parallel. For this reason, it is unclear which learning was inappropriate when punished. For example, it is not easy to distinguish whether the internal process was inappropriately switched or what a content word refers to was learned incorrectly. Furthermore, because participants are not instructed on what factors of the response should be evaluated, they may evaluate what the robot does not learn, such as response timing.

### C. Scaffolding Observed in Experiment 1

In the following, we will distinguish participants' utterances into two types [8]. One is a situation-matched utterance, called Yes-utterance. The other is an utterance inconsistent with the situation, called No-utterance. For example, when a participant says "This is an apple." or "Is this an apple?" while showing an apple, they are called Yes-utterances. However, if he/she gives the same utterances while showing a banana, they are called No-utterances.

14 undergraduates and graduates (10 males and 4 females) of engineering major participated in this experiment. While seven participants judged that learning was completed in less than 55 utterances (short duration group: white rows in Table I), the remaining seven subjects judged the completion with more than 99 utterances (long duration group: gray rows). Table I lists the total number of utterances, the number of Yes- and No-utterances including *yo*, *ne*, and *ka* respectively by each participant.

Participants in the long duration group changed the relative frequency of No-utterances in stages, which can be regarded as a kind of scaffolding. The dots in Fig. 3(a) show a typical stepwise change in the frequency of No-utterances, which were observed in *ne* utterances of Participant P2. The horizontal axis of the graph is the number of utterances, and the vertical axis on the left side is the difference between the cumulative numbers of Yes- and No-utterances with *ne* [5]. A Yes-utterance is plotted on the upper right of the previous dot, and a No-utterance lower right, that is, the y coordinate of the subsequent utterance is incremented by 1 if it is a Yes-utterance and decremented by 1 if it is a No-utterance. A blue circle designates that the utterance received a positive reward and a red triangle a negative reward.

---
[5] The number of dots in the graphs is less than the number of utterances. This is because these graphs are plotted only of utterances including a specific sentence-final particle, and the horizontal axis is the total number of utterances.

TABLE I. NUMBER OF UTTERANCES BY EACH PARTICIPANT.

| Partici-pants | Total # of U. | yo Yes-U. | No-U. | ne Yes-U. | No-U. | ka Yes-U. | No-U. |
|---|---|---|---|---|---|---|---|
| P1 | 195 | 49 | 15 | 32 | 18 | 34 | 47 |
| P2 | 148 | 16 | 5 | 55 | 35 | 22 | 15 |
| P3 | 164 | 29 | 21 | 48 | 35 | 16 | 15 |
| P4 | 27 | 4 | 3 | 7 | 2 | 6 | 5 |
| P5 | 21 | 6 | 0 | 7 | 3 | 3 | 2 |
| P6 | 53 | 5 | 1 | 18 | 12 | 11 | 6 |
| P7 | 158 | 13 | 3 | 75 | 47 | 11 | 9 |
| P8 | 195 | 50 | 8 | 42 | 32 | 34 | 29 |
| P9 | 100 | 26 | 16 | 20 | 14 | 15 | 9 |
| P10 | 115 | 23 | 0 | 38 | 3 | 32 | 19 |
| P11 | 11 | 3 | 0 | 4 | 0 | 2 | 2 |
| P12 | 55 | 11 | 0 | 21 | 0 | 11 | 12 |
| P13 | 42 | 10 | 0 | 16 | 0 | 15 | 1 |
| P14 | 33 | 13 | 0 | 11 | 0 | 9 | 0 |

The first blue dot at coordinates $(2, 1)$ shows that the second utterance was the first utterance includes *ne*, and it was a Yes-utterance received a positive reward. The first red triangle at (44, 15) represents that the first No-utterance with *ne* was issued here, and it got a negative reward. The next red triangle at (45, 16) tells that it was a Yes-utterance received a negative reward. These blue and red dots exhibit that in the beginning, only Yes-utterances were made, in the middle stage many No-utterances were given, and in the last stage, Yes-utterances became dominant again. Fig. 3(b) shows another typical example of a stepwise change observed in *yo* utterances of Participant P9.

In addition, the green line in Fig. 3(a) shows the transition of the selection probability of 'recall & comparison' processing for particle *ne*. The vertical axis on the right side represents the selection probability. The transition graph revealed that in the first stage, learning progressed so that 'memorization' is activated for *ne*, then when No-utterances started to be given, it turned out that 'memorization' is inappropriate and the robot gradually learned to select 'recall & comparison' instead. Fig. 3(b) shows a similar tendency.

Participants in the short duration group did not change the yes/no ratio by stages, because 1) they tended to use Yes-utterances mainly for *yo* and *ne*, and 2) they tended to use No-utterances from the beginning for *ka* to confirm the progress of learning.

### D. Results of DA Learning in Experiment 1

Table II lists the results of DA learning by each participant. The numerical value in each cell is the probability at which each action is selected at the end of the experiment. For example, in Participant P1, the rate at which memorization is selected for *yo* was 0.003, the rate of no speech response for *yo* was 0.989, the rate at which recall and comparison are selected for *ne* was 0.018, the rate of *nee* response for *ne* was 0.947, etc.

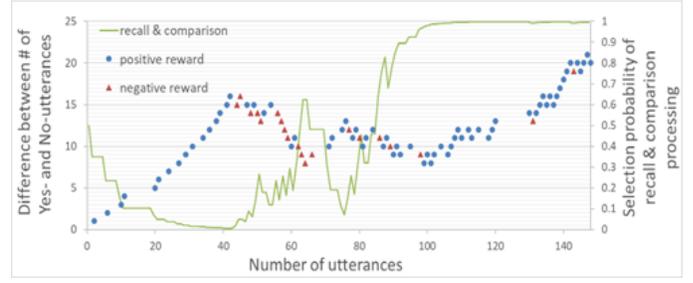

(a) *Ne* utterances of Participant P2.

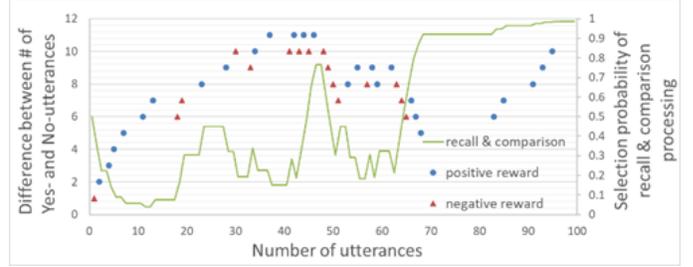

(b) *Yo* utterances of Participant P9.

Fig. 3. Typical examples of a stepwise change in the frequency of No-utterances and a corresponding learning curve.

TABLE II. RESULTS OF DA LEARNING BY EACH PARTICIPANT. THE NUMERICAL VALUE IN EACH CELL IS THE RATE AT WHICH EACH ACTION IS SELECTED AT THE END OF THE EXPERIMENT.

| Par-tici-pants | Total # of U. | yo Mem-orize | none | ne Re-call | nee | ka Re-call | none |
|---|---|---|---|---|---|---|---|
| P1 | 195 | 0.003 | 0.989 | 0.018 | 0.947 | 0.992 | 0.989 |
| P2 | 148 | 0.133 | 0.945 | 0.996 | 0.004 | 0.999 | 0.999 |
| P3 | 164 | 0.002 | 0.999 | 0.998 | 0.999 | 0.994 | 0.993 |
| P4 | 27 | 0.667 | 0.667 | 0.203 | 0.913 | 0.900 | 0.960 |
| P5 | 21 | 0.960 | 0.308 | 0.581 | 0.759 | 0.928 | 0.928 |
| P6 | 53 | 0.901 | 0.099 | 0.030 | 0.965 | 0.107 | 0.025 |
| P7 | 158 | 0.518 | 0.995 | 0.998 | 0.016 | 0.993 | 0.993 |
| P8 | 195 | 0.907 | 0.999 | 0.984 | 0.703 | 0.996 | 0.999 |
| P9 | 100 | 0.013 | 0.993 | 0.928 | 0.007 | 0.999 | 0.999 |
| P10 | 115 | 0.965 | 0.988 | 0.005 | 0.997 | 0.993 | 0.999 |
| P11 | 11 | 0.860 | 0.665 | 0.256 | 0.256 | 0.896 | 0.896 |
| P12 | 55 | 0.977 | 0.949 | 0.013 | 0.941 | 0.874 | 0.992 |
| P13 | 42 | 0.991 | 0.991 | 0.876 | 0.979 | 0.002 | 0.996 |
| P14 | 33 | 0.988 | 0.989 | 0.022 | 0.991 | 0.024 | 0.976 |

Initially, we assumed that learning would proceed as follows: internal processing for memorization is selected for *yo*, and internal processing for recall and comparison is selected for *ne* and *ka*. Depending on the characteristics of the participants' utterances, the result did not necessarily agree with the expectation, but they could be judged that reasonable learning was done.

For example, when participants gave many No-utterances with *yo* (See Table 1), the robot learned to select recall and comparison (Participants P1, P2, P3, and P9; red numbers in Table 2), and when participants gave several No-utterances with *yo*, the selection was at the chance level (Participants P4 and P7; blue numbers). These results were reasonable. When participants gave only or mostly Yes-utterances, memorization also is a reasonable choice even for questions, *ka*, the robot thus tended to nod to all yes-no questions (Participants P13 and P14; green numbers in Table 2). It is known that 2- and 3-years-olds have a tendency to answer yes to yes-no questions [13, 14], and it is interesting that the robot had a similar tendency.

In the early stage of the experiment, learning proceeded so that sympathy processing was selected for *ne* and the robot uttered *nee*. However, some participants started to utter No-utterances from the middle of the experiment, and in that case, it is inappropriate to respond with *nee*, so learning progressed so as not to speak *nee*. As a result, the robot no longer responded even to Yes-utterances with *nee* (Participants P2, P7, and P9; yellow cells in Table 2). This is because the robot was designed to switch the inner processing based only on the type of the sentence-final particles and it could not consider whether it was given a Yes-utterance or a No-utterance.

### E. Limitation of the Learning Model

The problem of *nee* responses to participants who make No-utterances can be solved if we redesign the necessary features for learning. However, we want robots themselves to notice this kind of problem, solve it, and continue learning. Specifically, as shown in Fig. 4, we want to make robots which can evolve through stages utilizing scaffolding provided by the partner.

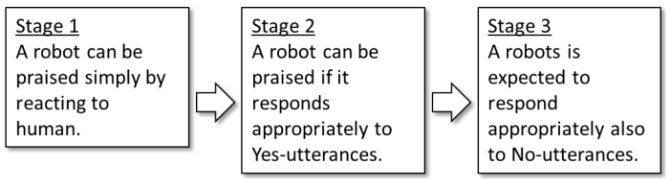

Fig. 4. Interaction that develops through stages.

Since the robot used in Experiment 1 was designed for the second stage, it could not keep up with the change when the way of interaction changed to the third stage. Then, if we design a robot for the third stage that can take complicated conditions into account, can it be used for the first stage and the second stage? It can learn also at early stages if it can spend enough time, but then people will not want to keep taking care of the robot. Fast learning according to the stage is necessary. To this end, we should implement some kind of attention mechanism that can handle the learning problem at each step as a simple one by narrowing down possibly related features to a small number.

## III. STEPWISE DA LEARNING

### A. Setting in Experiment 2

The procedure of Experiment 2, which was similar to that of Experiment 1, was as follows:

1. A participant talks to a robot while showing an apple or a banana. The participant's utterances are limited to 3 by 3 combinations of "apple/banana/looks tasty" and "*yo/ne/ka*".

2. The robot recognizes[6] the fruit and the speech, and makes a response. Expression from the robot is basically a movement of the neck (a nod, a head shake, or no motion), and in addition to this, the robot sometimes speaks *nee*.

3. The participant evaluates the response by pushing one of the + or − buttons.

Fig. 5 shows the experiment setting. Note that the participant has a piece of fruit in his hand, not putting it. Instructions for participants were the same as Experiment 1.

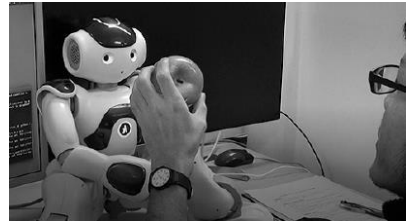

Fig. 5. Interaction setting in Experiment 2.

### B. Computational Model of Stepwise DA Learning

The structure of the proposed model is shown in Fig. 6. This model was designed to automatically detect the change from the 2nd stage to the 3rd stage and to continue learning according to the change. For this purpose, we added a "scaffolding detection" process, a "state space reconstruction" process, and a "memorization policy change" process to the first model described in Section II.*B*. They are shown in yellow blocks in Fig. 6, whose details are described later.

---

[6] In Experiment 2, the number of errors in image recognition and speech recognition was only one or two in the entire experiment. For this reason, manual corrections were not carried out.

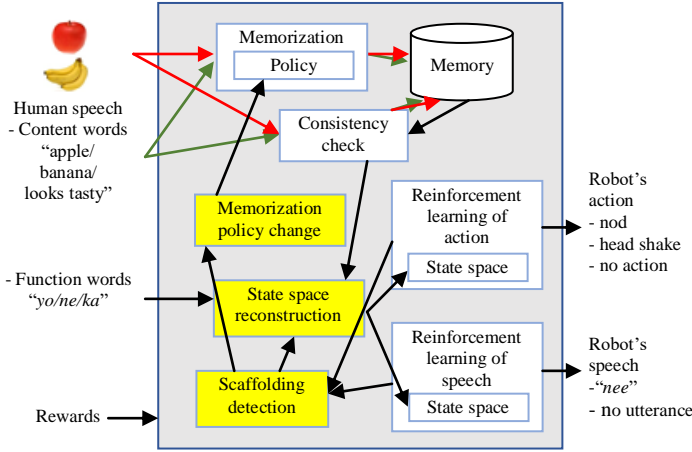

Fig.6. Structure of the stepwise DA learning system.

In order to focus on realizing the stepwise learning, two simplifications were applied to other parts: 1) Both "memorization" and "consistency check", which were called "memorization" and "recall & comparison" in the former model, are not switched, but both are always executed; 2) "Reinforcement learning (RL) of action and speech" were designed to learn external actions directly rather than producing external actions as a result of internal process learning.

"RL of action and speech" were implemented using equations (1) and (2) in Section II.B. $\alpha$ was set to 0.1 and $r$ was either $+1$ or $-1$. $\tau$ was set to 0.16. The state spaces of them start with three states corresponding to the three sentence-final particles but will be refined later. The behavior of the robot is either a nod, shaking head, or no action. While learning became simpler in that the behavior is learned directly, note that the choice of not moving the neck has added and it became difficult accordingly. The speech of the robot is *nee* or nothing.

We modeled the "memorization" process of content words as reinforcement learning (RL) here again. However, the design of the state and action has been changed. In this revised model, state $s$ in RL is the pair of a recognition result of fruits and one of the content words that was addressed, and action $a$ represents either the pair is consistent or not. Action-value function Q(s,a) is updated according to equation (1), and consistency is judged by the probability calculated by equation (2). Hence the recall threshold is unnecessary in this model.

The state spaces of "RL of action and speech" start with three states corresponding to the three sentence-final particles (left half of Fig. 7). After that, when a change in the learning stage is detected, the state space is divided according to the result of the "consistency check", that is, according to the distinction between Yes-utterance and No-utterance ("State space reconstruction"). If a negative reward is given despite selecting an action whose selection probability exceeds a threshold, it is judged that the learning phase has changed ("Scaffolding detection"), and the state space is divided. Note that only the state space that was used when receiving a negative reward is divided. For example, in the case where a negative reward is given despite the fact that the utterance which answered *nee* to *ne* was performed with high confidence, the state "*ne*" is divided into two states, "*ne* in Yes-utterances" and "*ne* in No-utterances" (right half of Fig. 7).

The first memorization policy was as follows: The pair of fruit in front of the robot and a content word that the robot heard is stored as a consistent pair regardless of a reward (equation (1)). Once the change of the stage was detected, the policy was changed as follows ("Memorization policy change"): if a positive reward is given, a fruit and a content word are memorized as a consistent pair, otherwise, as an inconsistent pair. When a reward is given for shaking head response, the polarity of the reward was reversed.

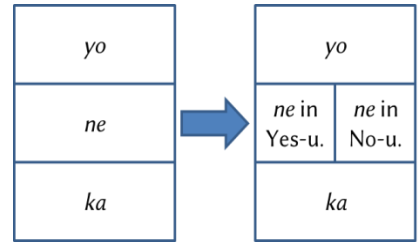

Fig.7. Example of a state space reconstruction.

### C. Scaffolding observed in Experiment 2

Twelve undergraduates and graduates (10 males and 2 females) of engineering major participated in this experiment. There was no overlap with the participants of Experiment 1.

In this experiment, the average length of interactions was short as about half compared with Experiment 1, and almost all utterances with *yo* and *ne* are Yes-utterances. Accordingly, no scaffolding similar to Experiment 1 was observed for *yo* and *ne*.

As for the sentence-final particle *ka*, there were three participants who gave a relatively large number of No-utterances (Participants S1, S9, and S12), and two of the three participants (Participants S1 and S12) used Yes-utterances frequently in the early stage and thus showed scaffolding similar to those shown in Fig. 3, which was a change from Stage 2 to Stage 3 (Fig. 4).

In addition, we found another type of scaffolding, tightening of evaluation, which means changing a reward from positive to negative for the same situation (Yes-utterance or No-utterance), the same sentence-final particle, and the same reaction of the robot. This corresponded to a change from Stage 1 to Stage 2 (Fig. 4). The emergence of tightening was seen in 5 participants, S1, S3, S4, S7, and S11, out of 12 participants. For S1 who gave a large number of utterances, we observed the emergence of tightening four times. These were changes in the participant's side according to the progress of robot learning.

For example, given a Yes-utterance with *ka* by S1, the robot responded with no action and a *nee* utterance six times. While S1 gave a positive reward on the first three occasions, gave a negative reward on the remaining three occasions in and after the middle stage.

## D. Results of Learning in Experiment 2

In the following, for ease of explanation, participants are divided into three groups according to the interaction duration, and the results of each group are described.

*1) Short Duration Group:* Five participants completed the experiment in less than 22 utterances (Participants S2, S6, S7, S8, and S11). Four of them did not give any No-utterances. One of them, S1, gave only one No-utterance. In the case of four participants who did not make No-utterances, the robot learned to nod to *yo* and *ka*. This was a reasonable result as the response in the second stage (See Fig. 4) because only the questions that should be answered "Yes" were addressed.

Regarding *ne*, the state space reconstruction occurred in 2 of 4 participants (S8 and S11), and when reconstruction occurred, learning proceeded to respond *nee* to Yes-utterances. Otherwise, regular responses were not learned. It is thought that this is because rewards were given focused on the motions. Specifically, a positive reward was given when the robot nodded, but a negative reward was given for no motion or a head shake. Utterances were considered only when there is no motion. The reward was negative when there is neither motion nor utterance, but the way of giving a reward was different for each participant when there was no motion and the voice response was *nee*.

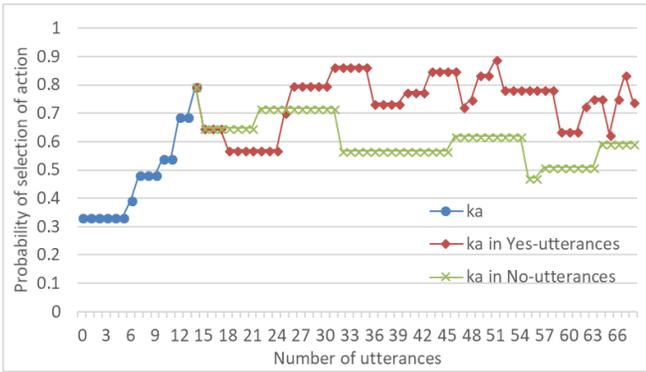

(a) A nod

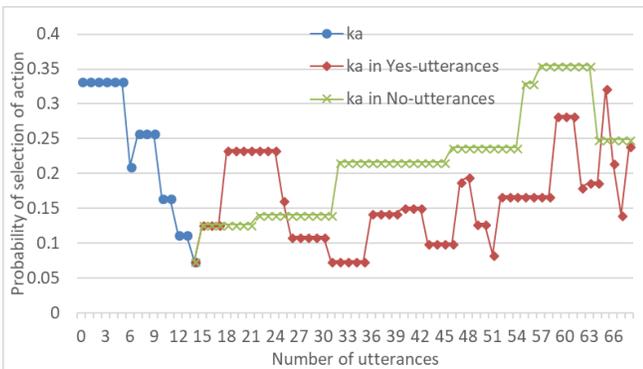

(b) A head shake

Fig.8. Transition of the probability of action selection for participant S12.

In the case of these five participants, the robot memorized each fruit and its name as a corresponding pair, and each fruit and "looks tasty" also as a corresponding pair.

In summary, if only Yes-utterances were given, it can be said that the robot could learn the proper response to sentence-final particles and the names of fruits quickly in most cases. This became possible because the robot has the ability of stepwise learning.

*2) Middle Duration Group:* Five participants (Participants S3, S4, S5, S10, and S12) completed the experiment with 32 to 68 utterances. They gave No-utterances from the beginning. Since this is different from the assumption, it did not match the initial memorization policy that assumes Yes-utterances, and as a result, the robot incorrectly memorized the image recognition result of an apple and the word "banana" as a corresponding pair and vice versa.

In the case of these five participants, learning for the sentence-final particles was also insufficient. Specifically, 4 out of 5 participants could not perform the state space reconstruction with action reinforcement learning for the sentence-final particle *ka*. Therefore, as a reaction for *ka*, learning progressed to do a nod or a head shake according to the frequency of Yes- and No-utterances. The robot could not learn to nod at Yes-utterances and to shake its head at No-utterances.

On the other hand, for Participant S12, the robot succeeded in performing the state space reconstruction for all the final particles in both the action reinforcement learning and speech reinforcement learning. The reason is that the number of No-utterances of the participant was 16, which was the largest among the middle duration group. However, learning of appropriate responses to Yes-utterances and No-utterances with the particle *ka* did not advance as shown in Fig. 8. Fig. 8 shows the transitions of the probability of action selection for S12. The horizontal axis shows the number of utterances and the vertical axis shows the selection probability of an action, a nod in Fig. 8(a) and a head shake in FIg. 8(b). The learning curves before the state space reconstruction are plotted with blue circles. The learning curves for Yes-utterances after the state space reconstruction are plotted with brown diamonds and the learning curves for No-utterances are plotted with green crosses. S12 did the scaffolding of increasing No-utterances in the second half but did not tighten the evaluation on the way. It might be said that 68 utterances without tightening were insufficient for progressing learning after space reconstruction.

*3) Long Duration Group:* Two participants (S1 and S9) completed the experiment with 94 and 105 utterances. They succeeded in letting the robot learn all sentence-final particles and content words almost correctly. In order to make a correct response to the sentence-final particle *ka* both for Yes-utterance (e.g. "Is this an apple?" while showing an apple) and No-utterance (e.g. "Is this an apple?" while showing a banana), it is necessary to perform the state space reconstruction in the state *ka* of the action reinforcement learning, and to learn to nod for Yes-utterances and to shake its head for No-utterances. Although they also gave No-utterances from the beginning, and the robot learned incorrect fruit names from No-utterances, eventually they let the robot learn correctly that an apple is NOT a "banana" etc. because they continued to interact with the robot enough.

obtain a definite conclusion, we observed that 1) the robot quickly learned to respond to DAs in most cases if the participants only spoke Yes-utterances, 2) in the case of participants who builds scaffolding differently from what we assumed, learning did not proceed quickly, and 3) the robot could learn to estimate DAs almost exactly if the participants kept interaction for a sufficiently long time even if the scaffolding was unexpected.

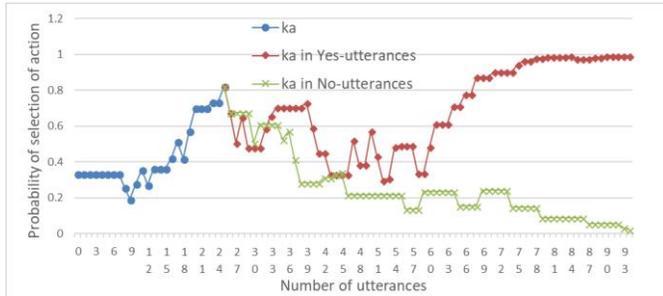

(a) A nod

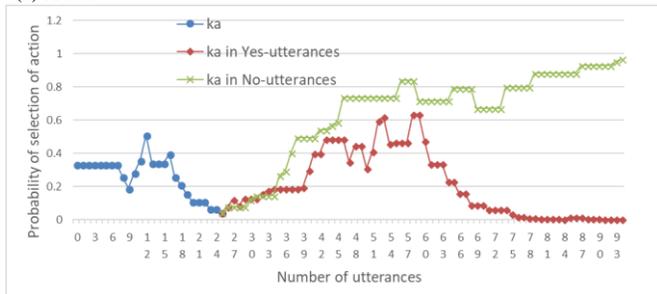

(b) A head shake

Fig.9. Transition of the probability of action selection for participant S1.

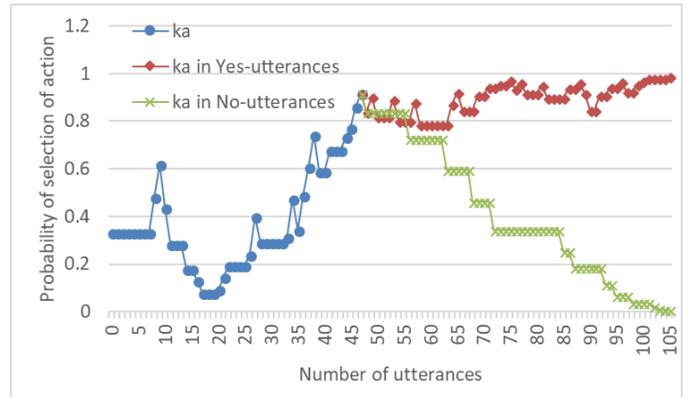

(a) A nod

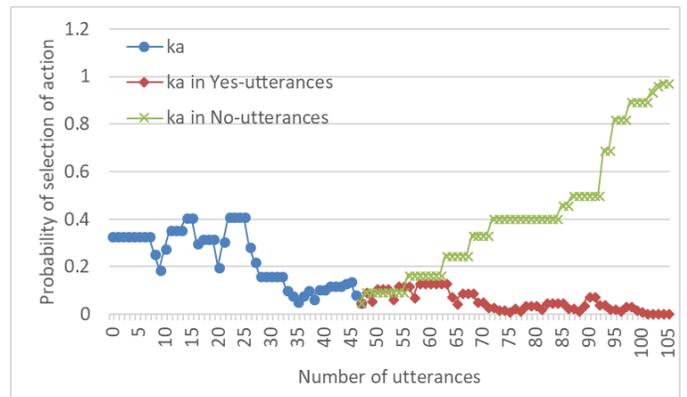

(b) A head shake

Fig.10. Transition of the probability of action selection for participant S9.

Participant S1 exhibited tightening of evaluation as mentioned in Section III.*C*. Fig. 9 demonstrates that the robot completely learned the actions for the sentence-final particle *ka*. Learning was almost completed even at the point of 68 utterances at which S12 in the middle duration group could not complete learning (cf. Fig. 8). Possible cause of the difference in learning speed between S1 and S12 is whether there was evaluation tightening or not.

The other participant S9 did not tighten the evaluation on the way. Learning thus was relatively slow (Fig. 10). Looking at the point of 68 utterances where S12 finished the interaction, we can see that the progress of learning of S9 and S12 was roughly equal. However, S9 succeeded in learning because the interaction was continued for a sufficiently long time.

IV. CONCLUSION AND FUTURE WORK

The purpose of this study was to verify that scaffolding takes place when a human teaches a robot, and to let the robot learn to estimate DAs and to make a response based on them step by step utilizing scaffolding. Experimental results demonstrated that participants who continued interaction for a sufficiently long time often gave scaffolding for the robot. Although the number of experiments is still insufficient to

We will try to increase the number of cases where the robot successfully utilizes the scaffolds. We believe that scaffolding and its utilization can be one of the breakthroughs to realize scalability because scaffolding works to reduce the size of the problem. Integration of supervised learning and reinforcement learning is another future task.

ACKNOWLEDGMENT

This work was supported by JSPS KAKENHI Grant Numbers 25330260, 18H05076.